\documentclass[pmlr,twocolumn,10pt]{jmlr} 

\usepackage{booktabs}
\usepackage{siunitx}

\usepackage{subfiles}
\usepackage{textcomp}
\usepackage{float}
\usepackage{graphicx}
\usepackage{amsmath}
\usepackage{multirow}
\usepackage{makecell}
\usepackage{array}
\usepackage{enumitem}

\usepackage{pifont}

\theorembodyfont{\upshape}
\theoremheaderfont{\scshape}
\theorempostheader{:}
\theoremsep{\newline}

\jmlryear{2022}
\jmlrworkshop{Conference on Health, Inference, and Learning (CHIL) 2022} 

\title[Uncertainty-Aware Text-to-Program for QA on Structured EHR]{Uncertainty-Aware Text-to-Program for \\Question Answering on Structured Electronic Health Records}

\author{%
\Name{Daeyoung Kim} \Email{daeyoung.k@kaist.ac.kr}\\
\Name{Seongsu Bae} \Email{seongsu@kaist.ac.kr}\\
\Name{Seungho Kim} \Email{shokim@kaist.ac.kr}\\
\Name{Edward Choi} \Email{edwardchoi@kaist.ac.kr}\\
\addr KAIST, Republic of Korea
}

\begin{document}

\maketitle

\begin{abstract}
Question Answering on Electronic Health Records (EHR-QA) has a significant impact on the healthcare domain, and it is being actively studied.
Previous research on structured EHR-QA focuses on converting natural language queries into query language such as SQL or SPARQL (NLQ2Query), so the problem scope is limited to pre-defined data types by the specific query language.
In order to expand the EHR-QA task beyond this limitation to handle multi-modal medical data and solve complex inference in the future, more primitive systemic language is needed.
In this paper, we design the program-based model (NLQ2Program) for EHR-QA as the first step towards the future direction.
We tackle MIMICSPARQL*, the graph-based EHR-QA dataset, via a program-based approach in a semi-supervised manner in order to overcome the absence of gold programs.
Without the gold program, our proposed model shows comparable performance to the previous state-of-the-art model, which is an NLQ2Query model (0.9\% gain).
In addition, for a reliable EHR-QA model, we apply the uncertainty decomposition method to measure the ambiguity in the input question.
We empirically confirmed data uncertainty is most indicative of the ambiguity in the input question.
\end{abstract}

\paragraph*{Data and Code Availability}
Our source code and dataset are available on the official repository\footnote{\url{https://github.com/cyc1am3n/text2program-for-ehr}}.


\section{Introduction}
\label{sec:intro}
Electronic health records (EHR) are composed of heterogeneous data (\textit{e.g.,} medical history, diagnoses, radiology images, and test results) generated after patients receive some form of medical service. EHRs are stored in databases with complex schemas such as in MIMIC-III~\citep{johnson2016mimic} or eICU~\citep{pollard2018eicu}.
It is hard for non-database experts to look up information or make decisions based on EHRs because they need to understand two things to obtain the information they want: the complex database structure and a query language such as SQL or SPARQL.
For example, to answer the question \textit{``what number of patients have been diagnosed with hyperglycemia?''}, one must generate a complex query such as \textit{``select count ( distinct patients.subject\_id ) from patients inner join admissions on patients.subject\_id = admissions.subject\_id where admissions.diagnosis = hyperglycemia''}.
Therefore, a real-time QA agent that can understand the structure of EHRs and make complex inferences would significantly lower the burden of medical personnel during decision making, patients seeking information, and researchers conducting medical research.

Recent works on EHR-QA with structured data (\textit{e.g.,} relational database or knowledge graph) have been focused on converting natural language questions (NLQ) into query languages such as SQL or SPARQL~\citep{wang2020text,park2020knowledge,bae2021uniqa} or into domain-specific forms~\citep{raghavan2021emrkbqa}.
However, because all previous works mentioned above rely on specific query languages, the problem scope is limited to pre-defined data types (\textit{e.g.,} string, int, timestamp) and operations.
To expand the EHR-QA task beyond the scope of a query language in order to conduct more complex inference and use multiple modalities (\textit{e.g.,} text, images, and signals), we require a program-based approach using atomic operations that are more primitive than those pre-defined by query languages.
For instance, given a sufficiently powerful vision component, the program-based approach can answer questions such as ``\textit{Did this patient have effusion in the left lung before he was admitted to the ICU?}".

\begin{figure}[t]
    \centering
    \includegraphics[width=0.9\columnwidth]{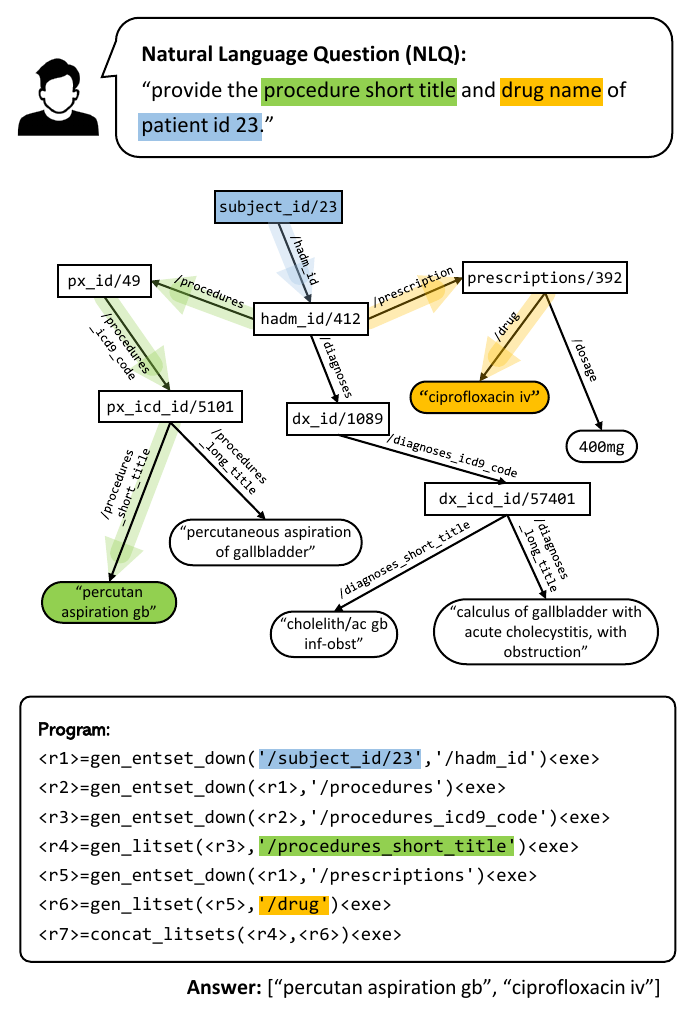}
    \caption{
    An illustrative example of our NLQ2Program approach for EHR question answering: a natural language question (NLQ), corresponding program traces over a knowledge graph, and its answer.
    }
    \label{fig:overview}
\end{figure}

Within the healthcare domain, the reliability of deep neural network models is crucial because incorrect decisions bring consequences such as ethical issues on human life or monetary cost~\citep{dusenberry2020analyzing}.
Likewise, EHR-QA models also need to be extremely reliable so that only correct answers are provided to the user, but the ambiguity in question due to lack of information or typos within the data make it challenging to train such reliable models.
For example, if the word \textit{``procedure''} is missing in the question in Figure~\ref{fig:overview}, the question becomes ambiguous since \textit{``short title''} could refer to either \textit{``procedure''} or \textit{``diagnoses''}, which increases uncertainty in data.
Therefore, measuring uncertainty to detect ambiguous questions helps make a reliable model, as the model can take appropriate actions such as asking the user for clarification.

For the first time, our work uses the natural language question-to-program (NLQ2Program) approach for EHR-QA.
Specifically, we tackle MIMICSPARQL*~\citep{park2020knowledge}, an EHR-QA dataset based on the open-source EHR data MIMIC-III.
Since MIMICSPARQL* consists of pairs of a natural language question (NLQ) and a corresponding SPARQL query, all previous studies tackled this dataset by translating NLQ to either SQL or SPARQL queries with varying degrees of success.
However, as stated above, we must venture beyond using a pre-defined query language such as SQL and SPARQL in order to handle multi-modal medical data and solve complex inference tasks in the future.
Therefore we tackle MIMICSPARQL* via an NLQ2Program approach in a semi-supervised manner, in order to overcome the fact that there is no ground truth program given for each NLQ to train the model with.
Our proposed model showed comparable performance to state-of-the-art NLQ2SQL or NLQ2SPARQL models that use the ground truth data.
Also, we propose a method for measuring the ambiguity of input questions with insufficient information using the ensemble-based uncertainty decomposition for each program token generated by the EHR-QA model.
We empirically demonstrate the effectiveness of using uncertainty decomposition to discern ambiguous questions, by evaluating MIMICSPARQL*'s test questions, where each question's ambiguity was manually annotated.

The contributions we make in this paper can be summarized as follows:
\begin{itemize}
    \item It is the first attempt at designing an NLQ2Program model that uses programs composed of various atomic operations for an EHR-QA task. Without ground truth programs, we obtained results on MIMICSPARQL*, the most recent EHR-QA dataset, comparable to the NLQ2SQL and NLQ2SPARQL SOTA models that use ground truth queries (0.9\% improvement).
    \item We generated a dataset to solve the problem without gold programs. We make it publicly available along with an interpreter that can execute programs so others can further research on EHR-QA using this NLQ2Program model in the future.
    \item We apply the ensemble-based uncertainty decomposition method to measure the ambiguity in the input question. To our best knowledge, this is the first attempt to detect ambiguous input questions in the QA research area. We show the effectiveness of measuring ambiguity using data uncertainty.
\end{itemize}

\section{Related Works}
\subsection{QA on Electronic Health Record}

Question answering on electronic health records (EHR-QA) can be divided into two broad categories: unstructured QA and structured QA.
In the former case, most works focus on the machine reading comprehension task on free-formed text such as clinical case reports~\citep{vsuster2018clicr} and discharge summaries~\citep{pampari2018emrqa}.
In the latter case, depending on database types of structured EHR, it can be further classified into two subcategories: table-based QA and graph-based QA. 
In both subcategories, EHR-QA is treated as a translation task, converting a natural language question into a query language (\textit{i.e.,} SQL/SPARQL) or a domain-specific logical form.
\citet{wang2020text} first released MIMICSQL, a large-scale table-based EHR-QA dataset for the Question-to-SQL generation task in the healthcare domain, and also proposed a sequence-to-sequence (seq2seq) based model TREQS, which translates natural language questions to SQL queries (NLQ2SQL).
\citet{park2020knowledge} constructed a Question-to-SPARQL dataset and treated EHR-QA as a graph-based task by converting the original tables of MIMICSQL into a knowledge graph. Also, they empirically showed that NLQ2SPARQL outperforms NLQ2SQL for the same dataset and the same model architecture. 

Recently, \citet{raghavan2021emrkbqa} constructed a new large-scale question-logical form pair dataset (emrKBQA) for MIMIC-III, which reuses the same logical forms proposed in emrQA~\citep{pampari2018emrqa}, but it is not currently publicly available.
Moreover, in order to execute the logical forms in emrKBQA, they must be mapped to corresponding SQL queries in advance.
To overcome the limitation of a query language (\textit{i.e.,} bound by pre-defined operations and only capable of handling fixed data types), we use the NLQ2Program approach for EHR-QA where programs are composed of atomic operations.
Specifically, we develop our NLQ2Program approach while viewing the EHR data as a knowledge graph rather than relational tables, similar to \citet{park2020knowledge,bae2021uniqa}, where the graph-based approach outperformed the table-based approach.

\subsection{Program Based Approach for KBQA}\label{subsec:programbasedapproach}
There are recent works translating natural language questions into multi-step executable programs over Knowledge Base Question Answering (KBQA; Structured QA)~\citep{liang2017neural, saha2019complex, hua2020less}.
These studies usually tackle datasets that do not have gold programs such as CQA~\citep{saha2018complex} and WebQuestionSP ~\citep{yih2015semantic}.
Specifically, Complex Question Answering (CQA)~\citep{saha2018complex} is a large-scale QA dataset that contains complex questions involving multi-hop and aggregation questions (\textit{e.g.,} counting, intersection, comparison), which are similar to our main target dataset (\textit{i.e.,} MIMICSPARQL*).
To handle the absence of the gold program, previous works proposed reinforcement learning (RL) based approaches. 

RL-based approaches, however, face challenges caused by the large search space and sparse rewards.
In particular, these challenges are intensified in MIMICSPARQL*, where KB artifacts (entities, relation, literal) are not explicitly revealed in the question.
For example, for the NLQ in Figure~\ref{fig:overview}, the QA model must generate a program using \textit{`/procedure\_icd9\_code'} and \textit{`/prescriptions'} which were never mentioned in the NLQ.
Moreover, MIMICSPARQL*'s search space is much larger than CQA since questions typically require a longer chain of operations to complete a program.
Therefore instead of using RL, we train our model in a semi-supervised manner and compare our approach with NS-CQA~\citep{hua2020less}, the state-of-the-art model for CQA.

\subsection{Uncertainty in Language Generation}
As the uncertainty in program-based EHR-QA has not been discussed before, we found the uncertainty in language generation to be the most relevant work to ours.
Recent approaches focus on predictive uncertainty by measuring the probability~\citep{ott2018analyzing} and the entropy~\citep{xu2020understanding, xiao2021hallucination} of each token in the generated sequence by the model.
\citet{xiao2021hallucination} apply the deep ensemble method~\citep{lakshminarayanan2017simple} to decompose uncertainty into \textit{data uncertainty}, the intrinsic uncertainty associated with data, and \textit{model uncertainty}\footnote{Data and model uncertainty are also called aleatoric and epistemic uncertainty.}, which reflects the uncertainty in model weights~\citep{der2009aleatory,kendall2017uncertainties}.
These works analyze relationships between the uncertainty during decoding and the final output quality rather than analyzing the ambiguity in the input.
Recently, \citet{malinin2020uncertainty} utilize uncertainty for error detection and out-of-domain (OOD) input detection.
For the OOD input detection, they focus on model uncertainty, which can capture discrepancies between the train and test datasets.
However, our interest is to deal with insufficient information in the input question, which raises data uncertainty rather than model uncertainty.

\begin{figure*}[t]
    \centering
    \includegraphics[width=2.0\columnwidth]{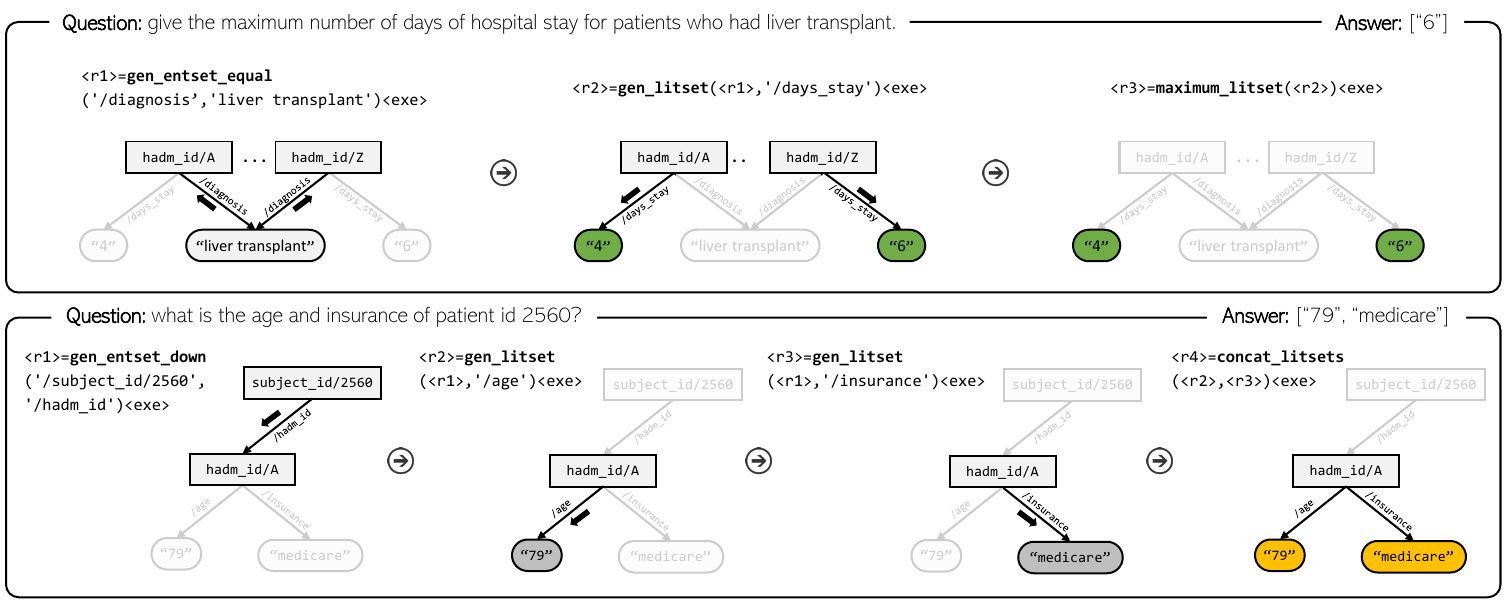}
    \caption{
    Illustrative examples of the natural language question (NLQ) and the corresponding program, composed of several predefined operations. To answer a single natural language question, we have to execute a series of atomic operations in sequence.
    }
    \label{fig:grammar}
    \vspace{-3mm}
\end{figure*}

\begin{table*}[!h]
\centering
\floatconts
{tab:grammar}
{\caption{Description of the custom-defined operations and return data types}}
{
\resizebox{\textwidth}{!}{%
\begin{tabular}{lll}
\toprule
\textbf{Operation}
& \textbf{Description}
& \textbf{Return Data Type}
\\ 
\midrule
gen\_entset\_down(\textit{entSet}, \textit{rel})
& the set of \textbf{object entities} associated with relation \textit{rel} for each subject entity in the \textit{entSet}
& \textit{entSet}
\\
gen\_entset\_up(\textit{rel}, \textit{entSet})
& the set of \textbf{subject entities} associated with relation \textit{rel} for each object entity in the \textit{entSet}
& \textit{entSet}
\\
gen\_litset(\textit{entSet}, \textit{rel})
& the set of \textbf{literal values} associated with the relation \textit{rel} for each subject entity of \textit{entSet}
& \textit{litSet}
\\
gen\_entset\_equal(\textit{rel}, \textit{lit})
& the set of subject entities which have literal value \textbf{equal to} \textit{lit} for relation \textit{rel}
& \textit{entSet}
\\
gen\_entset\_atleast(\textit{rel}, \textit{lit})
& the set of subject entities which have literal value of \textbf{at least} \textit{lit} for relation \textit{rel}
& \textit{entSet}
\\
gen\_entset\_atmost(\textit{rel}, \textit{lit})
& the set of subject entities which have literal value of \textbf{at most} \textit{lit} for relation \textit{rel}
& \textit{entSet}
\\
gen\_entset\_less(\textit{rel}, \textit{lit})
& the set of subject entities which have \textbf{smaller} literal value than \textit{lit} for relation \textit{rel}
& \textit{entSet}
\\
gen\_entset\_more(\textit{rel}, \textit{lit})
& the set of subject entities which have \textbf{greater} literal value than \textit{lit} for relation \textit{rel}
& \textit{entSet}
\\
count\_entset(\textit{entSet})
& the \textbf{number} of entities in \textit{entSet}
& \textit{int}
\\
intersect\_entsets(\textit{entSet$_1$}, \textit{entSet$_2$})
& the set of entities that \textbf{exist in common} in \textit{entSet$_1$} and \textit{entSet$_2$}
& \textit{entSet}
\\
maximum\_litset(\textit{litSet})
& the \textbf{largest value} in \textit{litSet}
& \textit{float}
\\
minimum\_litset(\textit{litSet})
& the \textbf{smallest value} in \textit{litSet}
& \textit{float}
\\
average\_litset(\textit{litSet})
& the \textbf{average value} of \textit{litSet}
& \textit{float}
\\
concat\_litsets(\textit{litSet$_1$}, \textit{litSet$_2$})
& the \textbf{combined list} of \textit{litSet$_1$} and \textit{litSet$_2$}
& \textit{litSets}
\\
\bottomrule
\end{tabular}
}}
\end{table*}

\section{Methodology}
\label{sec:methodology}
\subsection{Preliminary: Dataset}
In this work, we use a knowledge graph (KG) and questions from MIMICSPARQL*~\citep{park2020knowledge} consisting of 10,000 question-SPARQL pairs that cover 9 tables\footnote{Patients, Admissions, Diagnoses, Prescriptions, Procedures, Lab Results, Diagnosis Code Dictionary, Procedure Code Dictionary, Lab Code Dictionary} of MIMIC-III~\citep{johnson2016mimic}, an open-source ICU dataset.
Note that MIMICSPARQL* was derived from MIMICSQL~\citep{wang2020text}, a table-based EHR-QA dataset for MIMIC-III, consisting of 10,000 question-SQL pairs.
In other words, MIMICSPARQL* has the same question as MIMICSQL, but the ground truth queries and database format are different.
Also, note that each question in MIMICSPARQL* has two forms: template-based (machine-generated) form and natural (rephrased by medical domain experts) form.

\subsection{Grammar}
\label{sec:grammar}
We newly define a grammar that can effectively explore the KG of MIMICSPARQL*.
Since these are atomic operations that can be executed within KG, it is easy to handle multi-modality by expanding the grammar in the future.
This uses a total of 7 data types which are either KG artifacts or basic data types.
KG artifacts consist of \textit{entSet} (set of entities), \textit{rel} (relation), \textit{lit} (literal), \textit{litSet} (set of literals), \textit {litSets} (tuple of \textit{litSet}s), and basic data types consist of \textit{int} and \textit{float}.
Table~\ref{tab:grammar} and Figure~\ref{fig:grammar} describe the 14 operations we defined and their example.
We consider \citet{saha2019complex}'s work as our starting point, but we modify the set of operations to make them more suitable for complex EHR-QA.
For example, we add operations such as \textit{maximum\_litset}, \textit{minimum\_litset}, and \textit{average\_litset} since EHR-QA often requires calculations using numeric data found in the KG.

\subsection{Problem Formulation}
Our goal is to translate an EHR-related question into an executable program over KG.
Assume an underlying programming language $\mathcal{L}$.
Let us denote a given question by a sequence of tokens $Q=\{x_1,...,x_{\left\vert Q \right\vert}\}$ and the corresponding program $P\in \mathcal{L}$ can be represented as $P=\{y_1,...,y_{\left\vert P \right\vert}\}$.
Our model aims to maximize the conditional probability $p(P|Q)$.
Note that each question in MIMICSPARQL* has two forms, which are template-based (machine-generated) form and natural (rephrased by medical domain experts) form.
We define the former as $Q_T=\{x_1,...,x_{|Q_T|}\}$ and the latter as $Q_N=\{x'_1,...,x'_{|Q_N|}\}$.

\subsection{Synthetic Question-Program Generation}
\label{sec:synthetic_generation}
Since our method uses a custom set of operations (as described in Section~\ref{sec:grammar}), so our main obstacle to using NLQ2Program is the absence of gold programs (\textit{i.e.,} sequences of operations) for questions in MIMICSPARQL*.
We indirectly handle this problem by mass-generating pairs of MIMICSPARQL*-like questions and their corresponding programs.
Based on our preliminary analysis of the template-based questions in the MIMICSPARQL* train dataset, we first create a list of templates (\textit{e.g.,} what is the \textit{RELATION} of \textit{ENTITY}?) and question types (\textit{e.g.,} retrieve question).
Our analysis revealed that MIMICSPARQL* questions could be divided into total eight categories of question types.
Then we generate synthetic question $Q_{syn}$ and the corresponding program $P_{syn}$ pairs in a form similar to $Q_T$.
We sample program $P_{syn}$ and corresponding question $Q_{syn}$ by exploring KG while executing custom-defined operations for each of the eight question types.
Since KG schema of MIMICSPARQL* is complex, even if the same question type is given, the pattern of generated synthetic program sequence varies greatly depending on the required KG artifact (\textit{i.e.,} relation, entity).
Using this method, we generate 30,000 $(Q_{syn}, P_{syn})$ pairs for each type, and a total of 168,574 pairs are used after excluding duplicate questions or ones that already exist in MIMICSPARQL*.
Note that, it might be tempting to create gold programs by directly parsing the preexisting template questions $Q_T$, instead of creating synthetic questions and  programs.
This approach, however, has two major drawbacks: 1) it is complex; 2) Fragile to KG schema change.
Further details about the synthetic data generation process are presented in the Appendix~\ref{appendix:QG}.
Note that there are a good number of questions unlikely to be asked in the real-world setting, because values are sampled by randomly exploring the KG (\textit{e.g.,} \textit{how many patients whose language is engl and lab test value is 4.7k/ul?}).

\subsection{Semi-supervised Learning}
\paragraph{Obtaining NLQ \& pseudo-gold program pairs}
To acquire the pseudo-gold program $\tilde{P}$ for the corresponding natural language question $Q_N$ of MIMICSPARQL*, we introduce the following process:

\begin{enumerate}
  \item Train a supplementary sequence-to-sequence model $f_{syn}: Q_{syn} \mapsto P_{syn}$ using synthetic pairs (\textit{i.e.,} question and program $(Q_{syn}, P_{syn})$).
  
  \item Generate pseudo-gold programs $\tilde{P}$ by feeding $Q_{T}$ to the trained model $f_{syn}$ in order to obtain corresponding $(Q_{T}, \tilde{P})$ pairs.
  Here, $Q_{T}$ is used instead of $Q_{N}$ since $Q_{syn}$ and $Q_{T}$ are based on the same templates.

  \item Replace $Q_{T}$ with $Q_{N}$ to obtain pairs of natural-form questions and their corresponding pseudo-gold programs ($Q_{N}$, $\tilde{P}$).
\end{enumerate}

\paragraph{Training}
Train a sequence-to-sequence model $f_{\mathrm{N}}: {{Q_N} \mapsto \tilde{P}}$ with pairs of natural language question $Q_N$ and pseudo-gold program $\tilde{P}$. 
Note that the synthetic pairs $(Q_{syn}, P_{syn})$ are not used for training $f_{\mathrm{N}}$, and further experiments using synthetic pairs as pre-training data are shown in Appendix~\ref{appendix:pretrained_model}.

\subsection{Measuring Ambiguity of Question}
\label{sec:measure_ambiguity}
As mentioned above, we can use uncertainty in the output program to detect ambiguous questions that lack essential information (\textit{e.g.,} a user does not define the patient ID when asking for a patient's age) or include unseen values (\textit{e.g.,} typos).
Typically, uncertainty can be divided into data uncertainty and model uncertainty, where the former can be viewed as uncertainty measuring the noise inherent in given training data, and the latter as uncertainty regarding noise in the deep neural network parameters~\citep{chang2020data, dusenberry2020analyzing}.
Assuming that we can view ambiguous questions as inherent noise in the data (which the model cannot overcome by collecting more data, unlike model uncertainty), we aim to detect ambiguous input by measuring data uncertainty.
Following \citet{xiao2021hallucination, malinin2020uncertainty}, we adopt the ensemble-based uncertainty estimation method.

Given the question $Q=\{x_1,...,x_{\left\vert Q \right\vert}\}$ and the corresponding program $P\in \mathcal{L}$, we denote the context of the $i$-th program token $y_i$ as $c_i=\{x_1,...,x_{|Q|}, y_1, ..., y_{i-1}\}$, 
the prediction of each model in the ensemble of $M$ models as $\{p_m(y_i|c_i)\}_{m=1}^{M}$, and the aggregated prediction as  $p(y_i|c_i)={1\over{M}}\sum_{m=1}^{M}{p_m(y_i|c_i)}$.
Given context $c_i$, the entropy of $p_m(y_i|c_i)$ and $p(y_i|c_i)$ can be calculated as follows:
\begin{align*}
H_m(y_i|c_i) &= -\sum_{v\in\mathcal{V}} p_m(y_i=v|c_i)\log p_m(y_i=v|c_i) \\
H(y_i|c_i) &= -\sum_{v\in\mathcal{V}} p(y_i=v|c_i)\log p(y_i=v|c_i)
\end{align*}
where $\mathcal{V}$ is the whole vocabulary.
$H(y_i|c_i)$ represents the total uncertainty which is sum of data and model uncertainty. 
Then we can decompose $H(y_i|c_i)$ into data and model uncertainty as follows:
\begin{align*}
    u_{\text{data}} \left(y_{i}|c_{i}\right)
    &= \frac{1}{M} \sum_{m=1}^{M} H_m(y_i|c_i) \\
    u_{\text{model}} \left(y_{i}|c_{i}\right)
    &= H(y_i|c_i) - u_{\text{data}} \left(y_{i}|c_{i}\right)
\end{align*}
We assume the ambiguity of NLQ will raise the data uncertainty of a specific program token, not the entire program itself.
Note that $u_{\text{data}}(y_i|c_i)$ is calculated for every program token $y_i$.
We utilize the maximum value of the data uncertainty $u_{\text{data}}(y_i|c_i)$ for every program token $y_i$, instead of aggregating the $u_{data}(y_i|c_i)$ in a program-level manner~\citep{malinin2020uncertainty}.
Specifically, we determine if the input question is ambiguous using detector $g$ as follows:
\begin{align*}
    g(\mathbf{U} ; \tau)=
        \begin{cases}
        0 & \text { if }\mathrm{max}(\mathbf{U}) \leq \tau \\
        1 & \text { if }\mathrm{max}(\mathbf{U})>\tau
        \end{cases}
\end{align*}
where $\mathbf{U}=\{u_{\text{data}}(y_1|c_1), ..., u_{\text{data}}(y_{|P|}|c_{|P|})\}$ and specific threshold $\tau$.
We empirically show that this method can effectively detect ambiguous input questions.

\section{Experiments}
\subsection{Experiment Settings}
\subsubsection{Model Configurations}
Both the pseudo-gold program generating model $f_{syn}$ and the NLQ2Program model $f_N$ can be initialized with any sequence-to-sequence structure.
We choose T5-base~\citep{raffel2019exploring}, which is known to perform well in the natural language generation (NLG) field, for both models.
For comparison, we also experiment with UniQA~\citep{bae2021uniqa}, the state-of-the-art model in the MIMICSPARQL* dataset.
For the decoding strategy to generate program traces, we use beam search~\citep{wiseman2016sequence}.
Of the 8,000 samples from the MIMICSPARQL* training dataset, a total of 7,472 $(Q_N, \tilde{P})$ pairs that return the same execution result as the ground truth SPARQL query are used.
For an accurate evaluation, we use 949 of 1,000 samples from the MIMICSPARQL* test dataset after excluding samples whose ground truth SPARQL execution returns NULL, or whose questions and SPARQL queries do not match (\textit{e.g.,} the question adds the condition \textit{``less than 60 years of age''} while the ground truth query looks for \textit{``DEMOGRAPHIC.AGE $<$ 62''}).

\begin{table*}[t]
\centering
\floatconts
    {tab:base-exp}
    {\caption{
    Test results on MIMICSPARQL* with two different approaches: NLQ2SPARQL and NLQ2Program. 
    We report the mean and standard deviation of execution accuracy ($Acc_{EX}$) over 5 random seeds.
    }}
    {
    \resizebox{2.0\columnwidth}{!}{
    \begin{tabular}{ccccccccc}
    \toprule
    \multicolumn{1}{c}{\multirow{2}[1]{*}{\makecell{Recovery\\Technique}}}
    & \multicolumn{4}{c}{NLQ2SPARQL (w/ ground truth query)}
    & \multicolumn{4}{c}{NLQ2Program (w/o gold program)} \\
    \cmidrule(lr){2-5}
    \cmidrule(lr){6-9}
    & Seq2Seq
    & TREQS
    & UniQA
    & T5
    & NS-CQA (1\%)
    & NS-CQA (100\%)
    & Ours (UniQA)
    & Ours (T5)\\
    \midrule
    \ding{55}
    & 0.327 (0.043)
    & 0.699 (0.013)
    & 0.899 (0.010)
    & \textbf{0.905 (0.006)}
    & 0.203 (0.043)
    & 0.734 (0.087)
    & 0.860 (0.015)
    & 0.899 (0.005) \\ 
    \ding{51}
    & 0.338 (0.045)
    & 0.712 (0.011)
    & 0.939 (0.005)
    & 0.937 (0.006)
    & - 
    & - 
    & 0.920 (0.010)
    & \textbf{0.948 (0.006)} \\
    \bottomrule
    \end{tabular}
    }}
\end{table*}

\subsubsection{Baselines}
In the experiment, we compare our model against five baseline models as follows: Seq2Seq~\citep{luong2015effective}, TREQS~\citep{wang2020text}, UniQA~\citep{bae2021uniqa}, T5~\citep{raffel2019exploring} and NS-CQA~\citep{hua2020less}.
The first four are NLQ2SPARQL models using ground truth SPARQL queries, and the last is an NLQ2Program model which does not use gold programs.
Note that the T5 and UniQA models used as baselines adopt the NLQ2SPARQL approach, not NLQ2Program.
Among the previous program-based approaches mentioned in Section~\ref{subsec:programbasedapproach}, we choose NS-CQA as the baseline since it is the state-of-the-art model in the CQA~\citep{saha2018complex} dataset.
All models are trained with five random seeds, and we report the mean and standard deviation of performance.
The details of implementation are provided in Appendix~\ref{appendix:implementation_detail}.

\paragraph{Seq2Seq with Attention (NLQ2SPARQL)}\mbox{} \\
Seq2Seq with attention~\citep{luong2015effective} consists of a bidirectional LSTM encoder and an LSTM decoder.
Following the original paper, we apply the attention mechanism in this model. Note that this model cannot handle the out-of-vocabulary (OOV) tokens.
We denote the model as Seq2Seq.
\paragraph{TREQS (NLQ2SPARQL)}\mbox{} \\
TREQS \citep{wang2020text} is an LSTM-based encoder-to-decoder model using an attentive-copying mechanism and a recovery technique to handle the OOV problem.
\paragraph{UniQA (NLQ2SPARQL)}\mbox{} \\
UniQA~\citep{bae2021uniqa} is the state-of-the-art NLQ2Query model on MIMICSPARQL*. 
UniQA consists of a unified encoder-as-decoder architecture, which uses masked language modeling in the NLQ part and sequence-to-sequence modeling in the query part at the same time.
Following the original paper, we initialize UniQA with pre-trained BERT (12-layer, 768-hidden, 12-head)~\citep{devlin2018bert}.
\paragraph{T5 (NLQ2SPARQL)}\mbox{} \\
T5~\citep{raffel2019exploring} is a transformer-based encoder-to-decoder model which is pre-trained on a large corpus to convert every language problem into a text-to-text format.
We use T5-base model (12-layer, 768-hidden, 12-head) as mentioned above.
\paragraph{NS-CQA (NLQ2Program)}\mbox{} \\
NS-CQA \citep{hua2020less} is an LSTM-based encoder-to-decoder RL framework that obtains state-of-the-art performance on the CQA dataset.
It uses the copy mechanism and a masking method to reduce search space.
A memory buffer, which stores promising trials for calculating a bonus reward, is used to alleviate the sparse reward problem.
The model needs to be pre-trained by teacher forcing with pseudo-gold programs in order to mitigate the cold start problem.
We pre-train the model with two different data settings to study the effectiveness of the RL approach with restricted semi-supervision: (1) pre-train using all $(Q_N, \tilde{P})$ pairs, and (2) pre-train using only 1\% of all $(Q_N, \tilde{P})$ pairs (the same setting as \citep{hua2020less}). 
We then fine-tune the model by employing RL using all of the $Q_N$ and execution results of gold SPARQL query pairs.

\subsubsection{Evaluation Metric}
For comparing various models, three metrics are used in previous studies~\citep{wang2020text, park2020knowledge, bae2021uniqa}, which are \textit{Logical Form Accuracy} ($Acc_{LF}$), \textit{Execution Accuracy} ($Acc_{EX}$), and \textit{Structural Accuracy} ($Acc_{ST}$), to evaluate the generated queries (\textit{i.e.,} SQL, SPARQL).
However, $Acc_{LF}$ and $Acc_{ST}$ require gold programs since they compare the generated queries with the ground truth queries token by token.
Therefore, we only use \textit{Execution Accuracy} ($Acc_{EX}$), which measures the correctness of the answer retrieved by executing the generated program with the KG\footnote{This is why we excluded test samples whose answers are NULL, to minimize lucky guesses.}.

\subsubsection{Recovering Condition Values}
Following \citet{wang2020text}, we apply the recovery technique to handle inaccurately generated condition values that often contain complex medical terminology.
This technique replaces the condition values in the generated program with the most similar values that exist in the database.
For instance, the user may ask \textit{``how many patients had physical restrain status?''}, then one operation in the generated program could be \textit{ ``gen\_entset\_equal(`/diagnoses\_long\_title', `physical restrain status')''}. However, in the database, the value \textit{`physical restraints status'} exists, but \textit{`physical restrain status'} does not. In that case, the recovery technique replaces the incorrect condition value of the program to the correct one, thus making it executable.
In order to calculate the similarity between predicted values and existing ones, this technique uses ROUGE-L~\citep{lin2004rouge} score.

\subsection{Experiment Result}
As shown in Table~\ref{tab:base-exp}, despite the absence of gold programs, our NLQ2Program model is comparable with state-of-the-art NLQ2SPARQL models that require ground truth query data.
Also, we show that using NS-CQA's semi-supervised RL-based approach on MIMICSPARQL* is not effective when using only 1\% of the question and pseudo-gold program pairs for pre-training.

Note that the recovery technique is unnecessary for NS-CQA because, due to the decoding nature of NS-CQA, all KB artifacts (entity, relation, value) are copy-pasted from the NLQ to their appropriate locations after the program is generated.

Additionally, we conducted experiments regarding the effect of using the synthetic data introduced in Section~\ref{sec:synthetic_generation}, as pre-training data.
The detailed information is shown in Appendix~\ref{appendix:pretrained_model}.

\subsection{Ambiguous Question Detection}
In order to validate our method of measuring ambiguity using data uncertainty, we hand-annotated all MIMICSPARQL* test samples with the following labels.
According to the degree of ambiguity, we categorized the ambiguous questions into two types: (1) \textit{mildly ambiguous} (Mild) and (2) \textit{highly ambiguous} (High).
There are a total of 174 questions labeled as mildly ambiguous and 49 questions as highly ambiguous.
The rules of ambiguous question labeling are defined as follows:  
\begin{itemize}
    \item \textbf{Mildly Ambiguous (Mild)}: 
    We treat questions whose relations are not explicitly revealed in the NLQ, so it is hard to infer even with the condition value as mildly ambiguous questions.
    For instance, it is challenging to know whether the relation to \textit{``pneumococcal pneumonia''} is a \textit{/diagnoses\_long\_title} or a \textit{/diagnoses\_short\_title} for the question \textit{``provide the number of patients less than 83 years of age who were diagnosed with pneumococcal pneumonia.''}.
    However, if there is an ideal linker connected to the DB, the ideal linker knows that \textit{``pneumococcal pneumonia''} refers to a \textit{short\_title}. For this reason, we label these questions as mildly ambiguous.
    
    In addition, typos in the questions are unseen values for the EHR-QA model that also cause ambiguity, so questions with typos are also labeled as mildly ambiguous (\textit{e.g., what is the number of (dead $\rightarrow$) \textbf{cead} patients who had brain mass; intracranial hemorrhage?}).
    \item \textbf{Highly Ambiguous (High)}: 
    We consider a question as highly ambiguous when its NLQ is too vague that multiple correct programs can be generated.
    For the question \textit{``specify primary disease and icd9 code of patient id 18480''}, \textit{``icd9\_code''} could refer to either \textit{``procedure''} or \textit{``diagnosis''}.
    
    In addition, if the condition value in an NLQ corresponds to more than one relation, that NLQ is labeled as highly ambiguous.
    Note that even an ideal linker cannot find the exact relation of highly ambiguous questions.
    For example, the question \textit{``give the number of newborns who were born before the year 2168.''} is highly ambiguous since the condition value \textit{``newborn''} is related to both \textit{``/admission\_type''} and \textit{``/diagnosis''}.
    However, we label the NLQ as a mildly ambiguous question if there are implicit patterns of the phrase and relation pair, even though the NLQ can refer to multiple relations~(\textit{e.g.,} 96.8\% questions containing the phrase \textit{``emergency room''} are related to \textit{``/admission\_location''} instead of \textit{``/admission\_type''} in the corresponding programs).

\end{itemize}

\begin{table}[h]

\centering
\floatconts
{tab:ensemble-ambiguous}
{\caption{
Uncertainty results on MIMICSPARQL* with two ambiguity degrees: Mild \& High and High. 
We report AUPRC and AUROC for four types of uncertainty: $u_{data}$, $u_{model}$, $H$, and $H_{m}$. 
Note that $u_{data}$, $u_{model}$, and $H$ are calculated by the ensemble model consisting of 5 models, but $H_m$ is calculated by a single model. In case of $H_m$, we report the mean and standard deviation over 5 random seeds.
}}
{
\resizebox{0.5\textwidth}{!}{%
\begin{tabular}{cccccc}
\toprule
\multicolumn{2}{c}{Ambiguity}
& $u_{\text{data}}$
& $u_{\text{model}}$
& $H$
& $H_m$
\\
\midrule
\multirow{2}{*}{Mild \& High} & AUPRC & \textbf{0.476} & 0.445 & 0.446 & 0.427 (0.01) \\
& AUROC & 0.703 & \textbf{0.753} & 0.705 & 0.653 (0.01) \\
\hline
\multirow{2}{*}{High} 
& AUPRC & \textbf{0.198} & 0.140 & 0.157 & 0.158 (0.02) \\
& AUROC & \textbf{0.804} & 0.799 & 0.792 & 0.779 (0.01) \\
\bottomrule
\end{tabular}
}}
\end{table}




In this experiment, we utilize the EHQ-QA models trained after removing 27 question-program pairs whose $Q_T$ and $Q_N$ were unaligned, which could lead to incorrect uncertainty estimation.
We assess the detection capabilities of ambiguous questions based on data uncertainty $u_{\text{data}}$, model uncertainty $u_{\text{model}}$, total uncertainty $H$, and entropy $H_m$.
These types of uncertainty are calculated from all tokens in a single program sequence and we use the maximum value as the degree of uncertainty.
Note that $u_{\text{data}}$, $u_{\text{model}}$, and $H$ are calculated via ensemble model consisting of 5 models, but $H_m$ is calculated by a single model.
Performance is assessed via the area under a Precision-Recall (PR) curve and a receiver operating characteristic (ROC) curve.
The results in Table~\ref{tab:ensemble-ambiguous} show that using only data uncertainty detects ambiguous questions better than other metrics.
However, in the case of Mild \& High, AUROC of $u_\text{model}$ is higher than $u_\text{data}$.
This is due to the fact that there are some mildly ambiguous questions that have a correlation between the input question and the corresponding program, resulting in low $u_\text{data}$.
The EHR-QA model regards these questions as not ambiguous ones since the information is sufficient to generate a program by capturing the pattern of the question.
In addition, ambiguous questions cannot be perfectly detected because ambiguous questions are also included in the training data, and we only labeled ambiguous questions in the test samples.
We also compare the token-level and program-level methods of measuring uncertainty.
The results show the advantage of the token-level method.
Details of this experiment are available in Appendix~\ref{appendix:ambig_token_sentece}.

\begin{figure}[t]
    \centering
    \includegraphics[width=0.8\columnwidth]{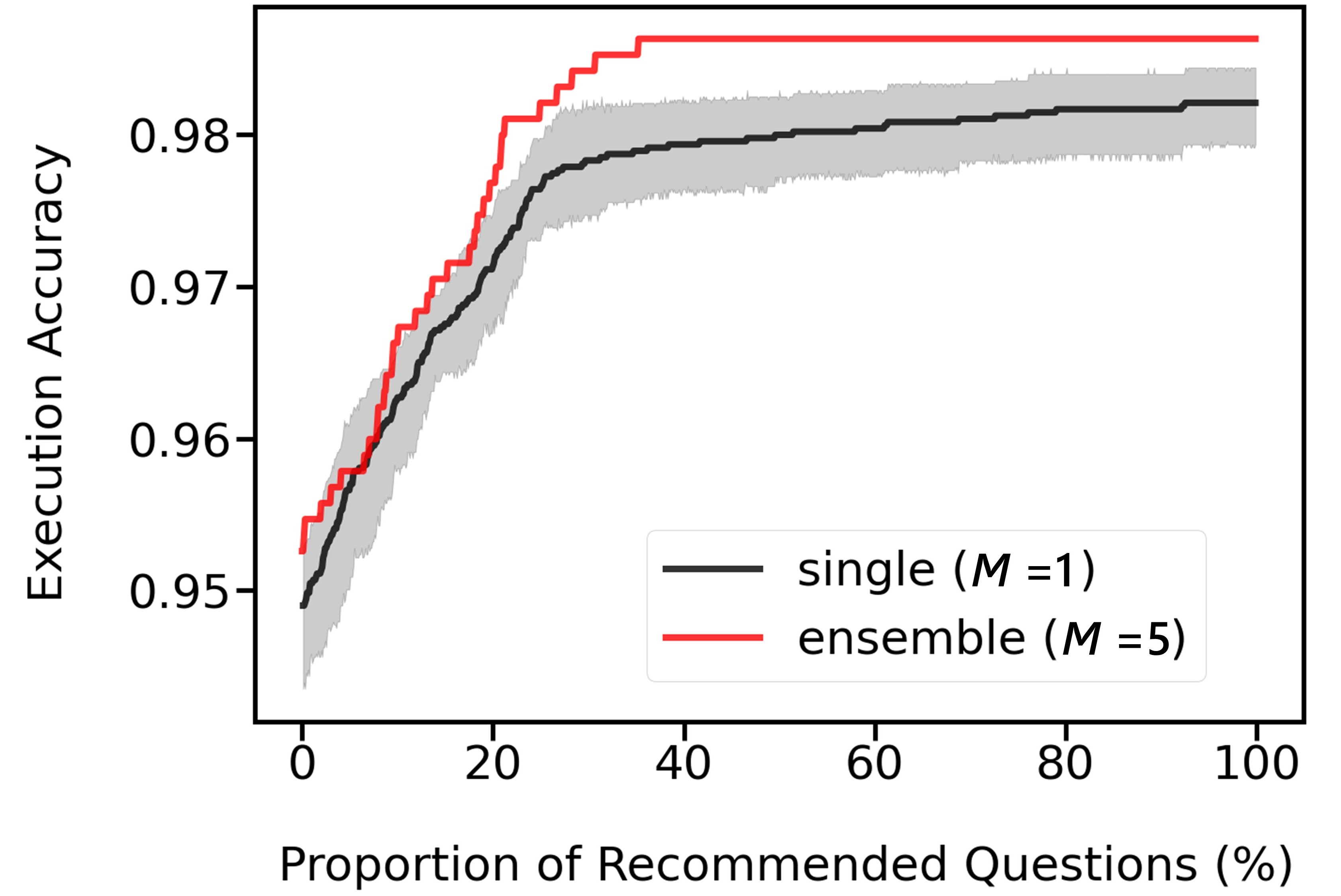}
    \caption{
    Test results of MIMICSPARQL* for five single models and an ensemble model.
    For the ensemble model, all tokens are generated by aggregating the prediction of all single models.
    The upper part of the shaded area presents the maximum execution accuracy of a single model, and the lower part shows minimum accuracy.
    }
    \label{fig:reanswer}
\end{figure}

To implement a more practical QA system, an interface is required to interact with the user, asking clarifying questions or allowing the user to modify the generated program.
Solving these issues, however, is beyond the scope of this work.
Instead, we build the system that gives users all five best beam hypotheses (\textit{i.e.,} programs) so users can select the appropriate candidate program when the ambiguity of the input question exceeds the specified threshold.
The results in Figure~\ref{fig:reanswer} show that the execution accuracy increases up to 0.986 with the number of recommendations.

\subsection{Qualitative Results}
\label{sec:qual_results}
In this section, we provide qualitative results to analyze generated programs along with token-level data uncertainty.
\subsubsection{Generated Programs for Ambiguous Questions}
\begin{figure}[t]
    \centering
    \includegraphics[width=1.0\columnwidth]{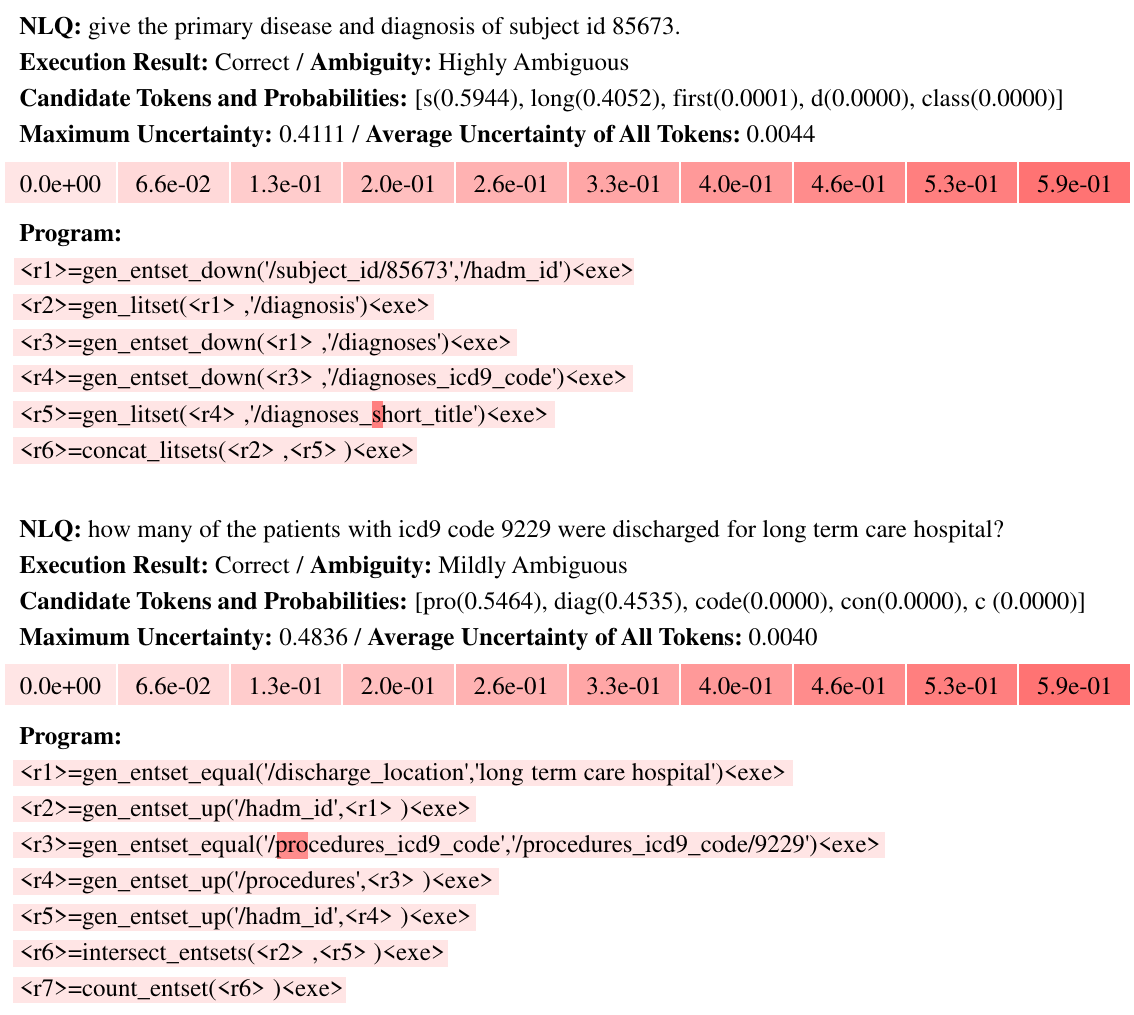}
    \caption{Qualitative results for ambiguous question. We visualize token-level data uncertainties of the generated program using the heatmap.}
    \label{fig:qual_sample1}
\end{figure}

As we expect, if the input question is ambiguous, high data uncertainty is measured.
The question in the first example in Figure~\ref{fig:qual_sample1} corresponds to a highly ambiguous question since whether the relation is either \textit{`long\_title'} or \textit{`short\_title'} is not specified.
It leads to the high uncertainty of the token \textit{`s'} whose data uncertainty is almost 100 times larger than the average data uncertainty of other tokens.
Note that if the model generates \textit{`long\_title'} rather than \textit{`short\_title'}, the execution result is incorrect but the program is semantically aligned with the question.
Likewise, in the second example, the question corresponds to a mildly ambiguous question since the \textit{`icd9 code 9229'} exists only in the \textit{`procedures.'}
However, it is hard to recognize this fact for the model, which does not have an ideal linker, so the data uncertainty is increased at the position of the token \textit{`pro.'}
These examples also demonstrate the max value of token-level uncertainty is an effective representative of the ambiguity.

\begin{figure}[t]
    \centering
    \includegraphics[width=1.0\columnwidth]{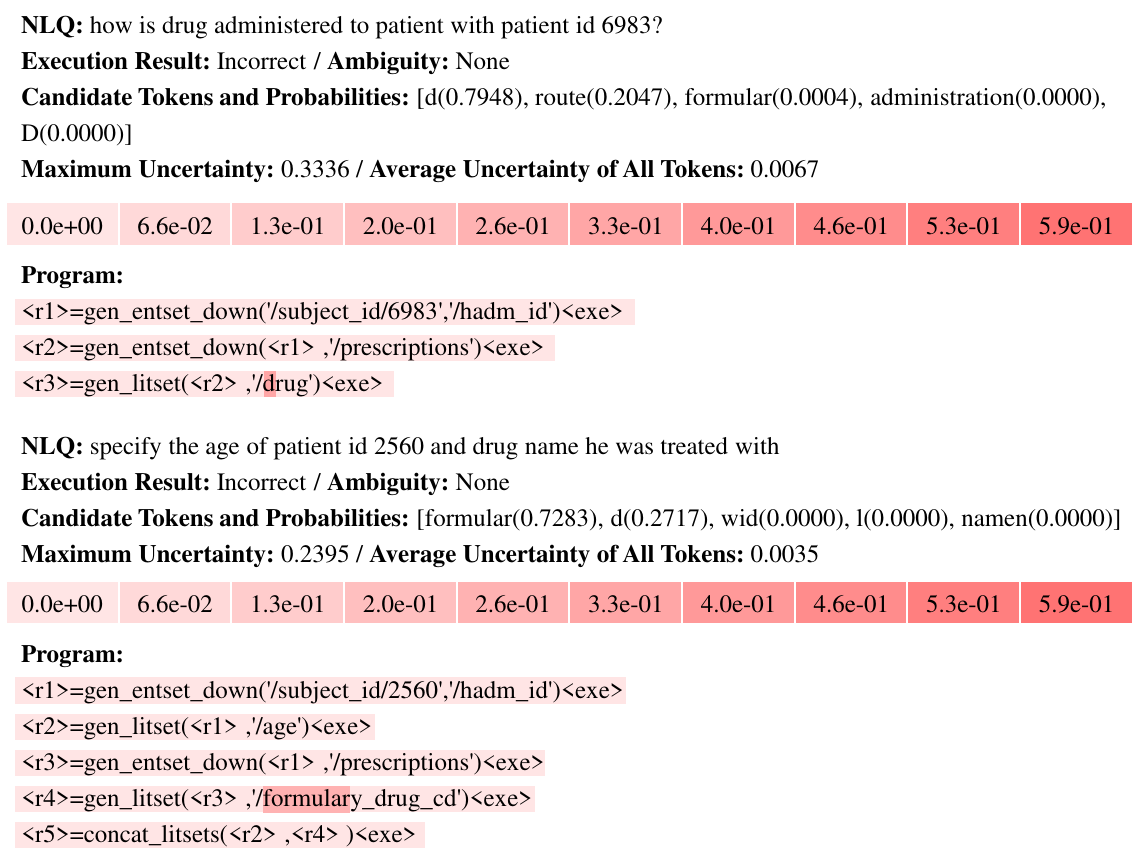}
    \caption{Qualitative results for failure cases. We visualize token-level data uncertainties of the generated program using the heatmap.}
    \label{fig:qual_sample2}
\end{figure}

\subsubsection{Failure Cases}
\label{sec:failure_cases}
There are some failure cases with high data uncertainty, but the NLQ is not ambiguous.
In the first example of Figure~\ref{fig:qual_sample2}, the token with the highest data uncertainty is \textit{``d,''}, a subset of an incorrect relation \textit{``/drug''} (drug name) that must be changed to \textit{``/route''} (route of administration).
Similarly, in the second example, the token with the highest data uncertainty is \textit{``formular,''}, a subset of relation \textit{``/formulary\_drug\_cd''} (drug code), which should be changed to \textit{``/drug''} (drug name). 
It can be seen that the data uncertainty does not always represent the ambiguity of the question.
However, when the model generates uncertain tokens, interaction with the user can still help improve the performance and reliability of the EHR-QA model.
Uncertainty in NLQ2Program is just the beginning, so more research is needed.

\section{Conclusion}
In this work, we designed an NLQ2Program methodology using atomic operations for EHR-QA task on MIMICSPARQL*.
We tackled the absence of gold programs via NLQ2Program approach in a semi-supervised manner.
Our proposed model showed comparable performance with the previous NLQ2SPARQL state-of-the-art model.
Moreover, we applied the ensemble-based uncertainty decomposition method to detect the ambiguous input question.
We showed the effectiveness of measuring ambiguity using data uncertainty.
Our further direction is to extend our methodology to handle multi-modal data on EHR and solve more complex questions.

\section*{Institutional Review Board (IRB)}
This research does not require IRB approval.


\acks{
This work was supported by Institute of Information \& Communications Technology Planning \& Evaluation (IITP) grant (No.2019-0-00075, Artificial Intelligence Graduate School Program(KAIST)) and National Research Foundation of Korea (NRF) grant (NRF-2020H1D3A2A03100945), funded by the Korea government (MSIT).
}

\bibliography{custom}

\clearpage
\appendix
\newpage
\section{Details for Synthetic Data Generation}
\label{appendix:QG}
Based on our preliminary analysis of the template-based questions in the MIMICSPARQL* machine-generated train dataset, we found there are eight basic templates as follows:
\begin{itemize}[leftmargin=5.5mm]
    \item what is \textit{RELATION} of \textit{ENTITY}?
    \item what is \textit{RELATION1} and \textit{RELATION2} of \textit{ENTITY}?
    \item what is \textit{RELATION1} of \textit{RELATION2} \textit{VALUE}?
    \item what is \textit{RELATION1} and \textit{RELATION2} of \textit{RELATION3} \textit{VALUE}?
    \item what is the number of \textit{ENTITY} whose \textit{RELATION} \textit{CONDITION} \textit{LITERAL}?
    \item what is the number of \textit{ENTITY} whose \textit{RELATION1} \textit{CONDITION1} \textit{LITERAL1} and \textit{RELATION2} \textit{CONDITION2} \textit{LITERAL2}?
    \item what is \textit{AGGR} \textit{RELATION1} of \textit{ENTITY} whose \textit{RELATION2} \textit{CONDITION} \textit{LITERAL}?
    \item what is \textit{AGGR} \textit{RELATION1} of \textit{ENTITY} whose \textit{RELATION2} \textit{CONDITION1} \textit{LITERAL1} and \textit{RELATION3} \textit{CONDITION2} \textit{LITERAL2}?
\end{itemize}
where \textit{CONDITION} corresponds to $=, >, <, \le, \ge$ and \textit{AGGR} represents $\min, \max$, and $\text{average}$.
For each template, we composed the operation set to be executed.
When we generate the synthetic question and corresponding synthetic program, the operation and corresponding argument to be selected for each step are arbitrarily determined.
Note that this simple technique is available for other KGs.

\section{Implementation Details}
\label{appendix:implementation_detail}
We implement our model and baseline models with PyTorch Lightning \footnote{\url{https://www.pytorchlightning.ai}} and HuggingFace's transformers\footnote{\url{https://huggingface.co/transformers/}}. In the case of TREQS and NS-CQA, we utilized the official code\footnote{\url{https://github.com/wangpinggl/TREQS}}\footnote{\url{https://github.com/DevinJake/NS-CQA/}} written by the origin authors.
Also in the case of UniQA, we manually implement model followed by descriptions in \citet{bae2021uniqa} since the official codes are not publicly available.
\subsection{Hyperparameters}
In order to make an accurate comparison with the baseline models, the Seq2Seq model and TREQS model were imported from \citet{park2020knowledge}, and hyperparameters were also imported with the same value. We trained our models on the following GPU environment: NVIDIA GeForce RTX-3090. Also, torch version is 1.7.0, and CUDA version is 11.1. Other hyperparameters are presented in Table \ref{tab:hparams}.

\begin{table*}[t]
\floatconts
{tab:hparams}
{\caption{Hyperparameters for training several models.}}
{
\resizebox{\textwidth}{!}{%
\begin{tabular}{cccccc}
\toprule
Hyperparameters & Seq2Seq & TREQS & UniQA & NS-CQA & Ours \\ \midrule
Hidden dimension & 256 & 256(enc) + 256(dec) & 768 & 128 & 768\\
Learning rate & \num{5e-4}  & \num{5e-4} & \num{3e-5} & \num{1e-3} (PT), \num{1e-4} (RL) & \num{1e-4}\\
LR Scheduler & \begin{tabular}[c]{@{}c@{}}StepLR(step size = 2,\\ step decay = 0.8)\end{tabular} & \begin{tabular}[c]{@{}c@{}}StepLR(step size = 2,\\ step decay = 0.8)\end{tabular} & Linear decay & Linear decay & Linear decay \\
Batch size & 16 & 64 & 18 & 32 (PT), 8 (RL) & 18\\
Epochs & 20 & 20 & 100 (w/ early stop) & 100 (PT), 30 (RL) (w/ early stop) & 100 (w/ early stop) \\
Seed & 1, 12, 123, 1234, 42 & 1, 12, 123, 1234, 42 & 1, 12, 123, 1234, 42  & 1, 12, 123, 1234, 42 & 1, 12, 123, 1234, 42\\
Beam size  & - & 5 & 5 & -  & 5\\
\bottomrule
\end{tabular}
}}
\end{table*}

\section{Performance Variance by Pre-trained Model}
\label{appendix:pretrained_model}
In Section~\ref{sec:synthetic_generation}, we introduced our synthetic data generation method via preliminary analysis of template-based questions in MIMICSPARQL* training set, and generated 168,574 synthetic question-program pairs.
With this large volume of pairs, we can utilize them as source data for further pre-training a model to improve its final performance.
Therefore, we investigate the utility of synthetic pairs for pre-training with three different types of models.
As shown in Table~\ref{tab:pretrain_model}, we can observe that three models, first initialized with BERT, improve after further pre-training with synthetic pairs.
However, our synthetic data cannot give performance gain because T5 is originally pre-trained with massive data so that our synthetic data cannot give model performance gain with the same quality as the original corpus of T5-base.

\begin{table}[h]
\floatconts
{tab:pretrain_model}
{\caption{
Test results of four pre-trained models on MIMICSPARQL* depending on whether each model utilizes synthetic data ($P_{syn}$, $Q_{syn}$) or not. 
We report mean and standard deviation of execution accuracy Acc$_{EX}$ over five random seeds.
}}
{
\resizebox{0.5\textwidth}{!}{%
\begin{tabular}{cccccc}
    \toprule
    \multicolumn{1}{c}{\multirow{2}[1]{*}{\makecell{Recovery Technique}}}
    & \multicolumn{1}{c}{\multirow{2}[1]{*}{\makecell{Synthetic Data \\ (P$_{syn}$, Q$_{syn}$)}}}
    & E-as-D
    & UniQA
    & BERT2BERT
    & T5-base \\
    &
    & (109M)
    & (109M)
    & (130M)
    & (220M) \\
    \midrule
    \ding{55}
    & \ding{55}
    & 0.877 (0.013)
    & 0.860 (0.015)
    & 0.854 (0.006)
    & \textbf{0.899 (0.005)} \\
    \ding{55}
    & \ding{51}
    & \textbf{0.882 (0.003)}
    & \textbf{0.896 (0.009)}
    & \textbf{0.893 (0.003)}
    & 0.896 (0.006) \\
    \ding{51}
    & \ding{55}
    & 0.939 (0.009)
    & 0.920 (0.010)
    & 0.913 (0.009)
    & \textbf{0.948 (0.006)} \\
    \ding{51}
    & \ding{51}
    & \textbf{0.944 (0.008)}
    & \textbf{0.938 (0.006)}
    & \textbf{0.940 (0.004)}
    & 0.944 (0.005) \\
\bottomrule
\end{tabular}
}}

\end{table}

\section{Token-level vs. Program-level Uncertainty Measuring}
\label{appendix:ambig_token_sentece}
We conduct the additional experiment to compare the token-level uncertainty measuring method with the program-level uncertainty measuring method.
Following \citet{malinin2020uncertainty}, we utilize the import weighting method using all beam hypotheses.
We use the normalizing factor for import weighting as 1 followed by the original paper.
We denote the data uncertainty, the model uncertainty, and the total uncertainty in the program-level as $U_{data}$, $U_{model}$, and $U_{total}$ respectively. 
Specially, $U_{data}$ is obtained by subtract $U_{model}$ from $U_{total}$.
\begin{table}[h]
\centering
\floatconts
{tab:sentence-ensemble-ambiguous}
{\caption{Token/Program-level uncertainty results for two different ambiguity degrees: Mild \& High and High. We report AUPRC and AUROC as evaluation metrics.}}
{
\resizebox{0.5\textwidth}{!}{%
\begin{tabular}{cccccccc}
\toprule
\multicolumn{2}{c}{\multirow{2}[1]{*}{Ambiguity}}
& \multicolumn{3}{c}{Token-level}
& \multicolumn{3}{c}{Program-level}
\\
\cmidrule(lr){3-5}
\cmidrule(lr){6-8}
& 
& $u_{\text{data}}$
& $u_{\text{model}}$
& $H$
& $U_{\text{data}}$
& $U_{\text{model}}$
& $U_{\text{total}}$
\\
\midrule
\multirow{2}{*}{Mild \& High} & AUPRC & \textbf{0.476} & 0.445 & 0.446 & 0.360 & 0.305 & 0.329 \\
& AUROC & 0.703 & \textbf{0.753} & 0.705 & 0.659 & 0.619 & 0.663 \\
\hline
\multirow{2}{*}{High} 
& AUPRC & \textbf{0.198} & 0.140 & 0.157 & 0.108 & 0.084 & 0.095\\
& AUROC & \textbf{0.804} & 0.799 & 0.792 & 0.704 & 0.609 & 0.673\\
\bottomrule
\end{tabular}
}}
\end{table}
The result shows that the token-level uncertainty measuring method is more effective than program-level when detecting ambiguous questions.
As we mentioned in Section~\ref{sec:measure_ambiguity}, we speculate uncertainty only increases at the missing information part as shown in Section~\ref{sec:qual_results}.
In addition, the result demonstrates the fact the data uncertainty is most indicative of the ambiguity in the input question consistently.

\end{document}